\documentclass{article}

\usepackage[accepted]{template/icml2024}
\usepackage{template/macro}

\usepackage[T1]{fontenc}
\usepackage[utf8]{inputenc}
\usepackage[english]{babel}

\usepackage{times}

\PassOptionsToPackage{hyphens}{url}
\usepackage{url}
\usepackage{hyperref}

\usepackage{graphicx}
\usepackage{caption}
\usepackage{tikz}
\usepackage{pgfplots}
\pgfplotsset{compat=1.17}

\usepackage{array}
\usepackage{booktabs}
\usepackage{microtype}
\usepackage{xfrac}
\usepackage{xcolor}
\usepackage{enumitem}
\setitemize{itemsep=0pt,parsep=0pt,topsep=0pt}
\setenumerate{itemsep=0pt,parsep=0pt,topsep=0pt}

\usepackage[normalem]{ulem}

\begin{document}








\title{Learning Associative Memories with Gradient Descent}
\author{
    {\bf Vivien Cabannes} \\ Meta AI
    \and 
    {\bf Berfin \c Sim\c sek} \\ Flatiron Institute 
    \and
    \bf{Alberto Bietti} \\ Flatiron Institute 
}
\date{}
\maketitle


\begin{abstract}
    
This work focuses on the training dynamics of one associative memory module storing outer products of token embeddings.
We reduce this problem to the study of a system of particles, which interact according to properties of the data distribution and correlations between embeddings.
Through theory and experiments, we provide several insights.
In overparameterized regimes, we obtain logarithmic growth of the ``classification margins.''
Yet, we show that imbalance in token frequencies and memory interferences due to correlated embeddings lead to oscillatory transitory regimes.
The oscillations are more pronounced with large step sizes, which can create benign loss spikes, although these learning rates speed up the dynamics and accelerate the asymptotic convergence.
In underparameterized regimes, we illustrate how the cross-entropy loss can lead to suboptimal memorization schemes. 
Finally, we assess the validity of our findings on small Transformer models.

\end{abstract}

\section{Introduction}

Modern machine learning often involves discrete data, whether it is labels in a classification problem, sequences of text tokens in language modeling, or sequences of discrete codes when dealing with other modalities.
In such settings it is common to consider cross-entropy objectives, and to embed each input and output token into high-dimensional embedding vectors.
Deep learning architectures consist in transforming the embedding vectors by a cascade of linear matrix multiplications together with non-linear operations.
This work aims at obtaining a fine-grained understanding of training a single such linear layer with the cross-entropy loss and fixed embeddings.
Indeed, one could then see the training of deep models as the joint training of multiple such associative memory models.
Although our setup admits a standard convex analysis treatment, we felt the need to provide a finer picture, more inline with behaviors observed when training large neural networks.

We consider~$N$ input tokens $x\in[N]$, each associated with some output $y = f^*(x)\in [M]$ for a deterministic function $f^*:[N]\to[M]$ and some number $M$ of classes.\footnote{We use the notation $[p] = \brace{1, 2, \dots, p}$.}
The input variable~$x$ is assumed to be drawn from a data distribution~$p(x)$.
The goal is to learn the input-output relationship~$f^*$ with a model of the form
\begin{equation}
    \label{eq:model}
    f_W(x) = \argmax_{y\in[M]} \scap{u_y}{W e_x}, \quad\text{with}\quad W\in\R^{d\times d}
\end{equation}
where $e_x,u_y\in\R^d$ are fixed input/ouput token embeddings with $d\geq 2$, and $W$ is a parameter to be learned.
The quality of a model~$f_W$ is typically measured through the 0-1 loss
\[
    \cL_{01}(f_W) = \E_{X,Y}[\ind{f_W(X)\neq Y}] = \sum p(x) \ind{f_W(x)\neq f^*(x)},
\]
while~$W$ is learned by optimizing a surrogate loss with gradient methods.
We will focus on the cross-entropy loss
\begin{align}
    \label{eq:obj}
    \cL(W) &= \E_{X,Y}\bracket{\ell(W; x, y)}
    \\\ell(W; x, y) &= \log(\sum_{z\in[M]} e^{\scap{u_{z}}{We_x}}) - \scap{u_y}{We_x}.
    \label{eq:loss}
\end{align}
This loss is also known as softmax, multinomial logistic or negative log-likelihood loss.

The model \eqref{eq:model} can be seen as an associative memory, which stores or ``memorizes'' pairwise associations $(x, y)$.
Associative memories originate in the neural computation literature, where they were used to model how the brain stores information \citep[see, e.g.,][]{Willshaw1969,Willshaw1970}, and solve algorithmic tasks with a machine learning perspective \citep[see, e.g.,][]{hopfield1982,hopfield1985}.
These models have gained in popularity recently, notably as candidates to explain the inner workings of some deep neural networks~\citep{geva2020transformer,Schlag2021LinearTA,ramsauer2021hopfield,bietti2023birth,cabannes2023scaling}.
This line of work motivates the thorough study of the training dynamics behind associative memory models.
This is the goal of our paper, formalized as Problem \ref{pb}.

\begin{problem}
    \label{pb}
    Understand the training dynamics of the associative memory model with the cross-entropy loss.
\end{problem}

Training dynamics on the logistic loss have been the subject of a vast line of work~\citep[see, e.g.,][]{soudry2018implicit,ji2019implicit,ji2021characterizing,lyu2019gradient,wu2023implicit}.
In contrast to these works, we focus on the specific structure arising from our associative memory setup.
In particular, we characterize phenomena observed empirically in deep learning, such as loss spikes and oscillations, which are often necessary for faster optimization.
While oscillatory behaviors have been studied in the literature on large learning rates and \emph{edge-of-stability}~\citep{cohen2020gradient,nakkiran2020learning,beugnot2022benefits,agarwala2022second,bartlett2023dynamics,chen2023beyond,rosenfeld2023outliers,wu2023implicit}, our setup provides a new perspective on the matter, involving correlated embeddings and imbalanced token frequencies.

\paragraph{Contributions.}
With the goal of resolving Problem~\ref{pb}, we make the following contributions:
\begin{itemize}
    \item We show that all gradient dynamics (i.e. stochastic or deterministic, continuous-time or discrete-time) reduce to a non-linear system of interacting particles.
    \item We solve the deterministic dynamics for orthogonal embeddings, where the memories do not interfere much.
    \item We illustrate typical training behaviors in the overparameterized case $d \geq N$ by solving the system with two-particles.
    In particular, we show that competition between memories can lead to benign oscillations and loss spikes, especially when considering large step-size, although large learning rates accelerate the asymptotic convergence toward robust solutions of the underlying classification problem.
    \item In limited capacity regimes ($d < N$), we illustrate precisely the deterministic dynamics for $d=M=2$, when $N \geq 2$ particles interact.
    We showcase how the competition between memories can ultimately erase most of them.
\end{itemize}
We complement our analysis with experiments, investigating small multi-layer Transformer models with our associative memory viewpoint and identifying similar behaviors to those pinpointed in the simpler models.

\section{The Many Faces of Problem~\ref{pb}}

This section focuses on the disambiguation of Problem~\ref{pb}.

\paragraph{What type of understanding.}
Although the training dynamics are the result of deterministic computations on a computer, which could be described exhaustively, the causal factors behind their behavior are often too numerous for us to fully comprehend, even for the simple model~\eqref{eq:model}.
To overcome this issue, one can abstract coarse quantities that play important roles in many training scenarios, such as (i) the dimension and geometry of embeddings; (ii) data distributional properties, such as imbalanced token frequencies and heavy tails; (iii) the optimization algorithms and their hyperparameters, particularly the learning rate.
We aim to highlight the effect of these factors, whose understanding would help predict the outcome of alternative training choices such as bigger learning rates, or different data curricula.
Levels of understanding vary from rigorous math on small models, to controlled experiments on more complex models, or insights from the training of large-scale models without many ablation studies.
This paper aims for a theoretical study in between these two extremes.

\paragraph{Which dynamics.}
We classify dynamics into five types, providing coarse and fine approximations of the dynamics used in practice to train neural networks.

{\em Gradient flow.}
The gradient flow dynamics consists in letting the weight matrix $W$ evolves according to the equation
\begin{equation}
    \label{eq:GF}
    \diff W = -\nabla\cL(W_t) \diff t.
\end{equation}
From initialization $W=W_0$, this deterministic evolution pushes $W_t$ towards the lower value of the ``potential'' $\cL$.
In our case, $\cL$ is convex, but does not always have a minimum as it might be minimized by a $s W_*$ for $s$ going to infinity.

{\em Gradient descent.}
Gradient descent is the discrete-time approximation of the gradient flow dynamics, namely
\begin{equation}
    \label{eq:GD}
    \Delta_t W = -\eta_t\nabla\cL(W_t),
\end{equation}
where $\Delta_t W = W_{t+1} - W_t$ and $\eta_t$ is a learning rate.
In the continuous case, the learning rate corresponds to a reparameterization of the time with $ds = \eta_t \diff t$.
However, in the discrete-time regime, the learning rate is an important parameter that does influence the dynamics.

{\em Stochastic gradient flow.}
In practice, following full batch dynamics, i.e., dynamics that involve processing all the data at all time to compute the gradient of $\cL(W)$, is quite costly and inconvenient.
In those cases, one can process a random subset of data instead to get a good estimate of $\nabla \cL(W)$.
This randomness can be modeled as a perturbation of the dynamics due to some random noise with zero mean.
For gradient flow, the stochasticity is naturally modeled with a Brownian motion $\cE_t$,
\begin{equation}
    \label{eq:SGF}
    \diff W = -\nabla\cL(W_t) \diff t + \sigma_t \diff \cE_t
\end{equation}
where $\sigma_t$ is the variance of the updates, i.e. of the gradient $\nabla \cL(W_t;B) := \sum_{(x,y)\in B} \nabla \ell(W_t; x, y)$, when considering random batches $B$ of data.

{\em Stochastic gradient descent.}
For discrete-time dynamics, stochastic gradient descent can be written as
\begin{equation}
    \label{eq:SGD}
    \Delta_t W = -\eta_t(\nabla\cL(W_t) + \epsilon_t),
\end{equation}
where $\epsilon_t$ is some random variable.
When the descent is ``unbiased'', this variable has zero mean.
Typically, 
\[
    \nabla\cL(W_t) + \epsilon_t = \sum_{(x,y)\in B} \nabla \ell(W_t;x,y),
\]
for some random mini-batch $B$ of data with $y=f^*(x)$.

{\em Practical descent.}
Practitioners often use variants of stochastic gradient descent that are known to perform well empirically.
These typically involve momentum in the descent, re-conditioning the gradient \citep{kingma2017adam}, and the addition of normalization layers in the architecture~\citep[see, e.g.,][]{ioffe2015batch,ba2016layer}.

This paper focuses on gradient flow and gradient descent, postponing the study of other dynamics for future work.
We discuss the consistency of these methods, i.e., if they are reaching the best possible performance, their asymptotic convergence behaviors, as well as finite-time behaviors.

\section{Memories as Interacting Particles}

The section reduces the training dynamics to a system of interacting particles, where the particles correspond to input-output associations. 

To simplify the analysis of our model, a few simple observations can be made.
First, one can identify matrices with two-dimensional tensors, highlighting the linearity of our model $\scap{u_j}{We_i} = \scap{W}{u_j\otimes e_i}$ in the tensor space $\R^d\otimes \R^d \cong \R^{d\times d}$.
Secondly, recall that the loss \eqref{eq:loss} corresponds to the negative log-likelihood
\[
    \ell(W; x, y) = - \log p_W(y|x), 
\]
of the probability $p_W$ whose conditionals are parameterized as a soft-max over the scores $\scap{e_x}{W u_y}$,
\begin{equation}
    \label{eq:model-prob}
    p_W(y|x) \propto \exp(\scap{W}{e_x\otimes u_y}). 
\end{equation}
The chain rule leads to the following formula,
\begin{equation}
    \label{eq:grad}
    \nabla \ell(W; x, y) 
    = \sum_{z\in[M]} p_W(z\,\vert\, x) (u_z - u_y) \otimes e_x.
\end{equation}
The gradient formula shows that the dynamics take place on the span of the $(u_j - u_k)\otimes e_i$ with $i\in[N]$ and $j,k\in[M]$ up to an affine shift due to initialization. 

The resulting training dynamics can be studied by tracking projections onto the input and output embeddings, or onto another family generating the tensor space $\R^d \otimes \R^d$, which leads to a system of particles with non-linear interactions.

\begin{theorem}[Particle system]
    \label{thm:inter}
    Define the particle $w_{ij}$,
    \begin{equation}
        \label{eq:particles}
        w_{ij} = \scap{W}{u_j\otimes e_i} = u_j^\top W e_i,
    \end{equation}
    as well as the constant correlation parameters
    \begin{equation}
        \label{eq:correlation}    
        \alpha_{ij} = \scap{e_i}{e_j}, \quad \beta_{ijk} = \scap{u_i}{u_j - u_k}.
    \end{equation}
    The projected gradient can be rewritten as
    \[
        \scap{\nabla \ell(W; x, y)}{u_j\otimes e_i}
        = \alpha_{ix}\sum_{z} \frac{\beta_{jzy}\exp(w_{xz})}{\sum_{k \in[M]} \exp(w_{xk})}.
    \]
    Hence, all variations of gradient dynamics, \eqref{eq:GF}, \eqref{eq:GD}, \eqref{eq:SGF} and \eqref{eq:SGD}, can be expressed as a (stochastic) system of interacting particles.
    For example, the gradient descent dynamics \eqref{eq:GD} is
    \begin{equation}
        \label{eq:interaction}
        \Delta_t w_{ij} = \eta_t \sum_{x} p(x) \alpha_{ix} \sum_z\frac{\beta_{jf^*(x)z} \exp(w_{xz})}{\sum_{k\in[M]} \exp(w_{xk})}.
    \end{equation}
    Similarly, the dynamics for the stochastic gradient descent consists in replacing $\sum_x p(x) $ by the summation over $x$ in a random mini-batch in \eqref{eq:interaction}.
\end{theorem}
\begin{proof}
    The proof follows directly from \eqref{eq:model-prob} and \eqref{eq:grad}.
\end{proof}

There are two reasons for interactions in this particle system; either the input embeddings are not orthogonal and the $\alpha$'s mix the particles, or there are more than two classes and $\beta$'s mix the particles.
Moreover, when the embeddings are not orthogonal, particles are not independent, since an increase of $w_{ij}$ changes $w_{ik}$, as soon as $u_i^\top u_k \neq 0$.
Note that multiple factors could lead to correlated embeddings, such as under-parameterization (viz., embeddings are necessarily correlated in low dimension), or semantic similarity in the case of trained embeddings~\citep[e.g.,][]{mikolov2013distributed}.

Interacting particle systems commonly arise in other machine learning settings, e.g., to describe parameters in the mean-field regime of two-layer networks~\citep{chizat2018global,mei2018mean,rotskoff2018trainability}, samples in certain approaches to generative modeling~\citep{liu2016stein,arbel2019maximum}, or both~\citep{domingo2021dual}. However, these systems typically involve particles as discretizations of an underlying measure evolution, while we make no such connection here. Our dynamics may also be seen as training the middle layer of a three-layer linear network, and infinite-width dynamics for related models have been studied in~\citep{jacot2021saddle, chizat2022infinite}. Yet, our focus is on the finite width ($d< \infty$) case, and we note that this suffices for optimal storage when~$d$ is sufficiently large compared to~$N$ and~$M$.

The particle $w_{xi}$ corresponds to the score assigned by~$W$ to the class~$i$ for the token~$x$.
Another set of sufficient statistics for the problem are the margins, which are defined by
\begin{equation}
    \label{eq:margin}
    m_i(x) = w_{xf^*(x)} - w_{xi} =  (u_{f^*(x)} - u_i)^\top We_x.
\end{equation} 
It corresponds to the difference between the scores assigned by the model \eqref{eq:model} to the classes $f^*(x)$ and $i$, for the input $x$.
When all the margins $(m_i(x))_i$ are positive, the token $x$ is classified correctly.

\section{Overparameterized Regimes}

This section focuses on the case where $N \leq d$ and the $(e_x)$ form a linearly independent family.
In this setting, the optimization of the convex loss \eqref{eq:obj} will ensure perfect accuracy for our model, i.e. $f_W=f^*$.

\subsection{Orthogonal Embeddings}

We first solve the case where the embedding families $(e_x)$ and $(u_y)$ are both orthogonal.
The orthogonality of the inputs implies $\alpha_{ix} = \ind{i=x} \norm{e_i}^2$, in which case \eqref{eq:interaction} shows that the gradient dynamics for $W$ decouple on the $\R^d \otimes e_i$.
In other terms, our model is implicitly fitting in parallel $N$ parameters, the $(We_i)_i$, of $N$ independent exponential families, the $p_W(y|x)$.
As a consequence, we can forget the context variable and fix a $x\in[N]$ for the remainder of the section. 
For simplicity, we assume $f^*(x) = 1$.

\paragraph{Binary classification.}
Let us consider the binary case first.
When $M=2$, the dynamics on $We_x$ evolves on the line $\R\cdot(u_1 - u_2)$, and is fully characterized by the margin
\[
    m_t = (u_1 - u_2)^\top W_t e_x.
\]
An algebraic manipulation of~\eqref{eq:interaction} shows that this scalar quantity evolves according to the dynamics
\[
    (1 + \exp(m_t)) \Delta_t m = c_x\eta_t,
\]
$c_x = p(x) \norm{e_x}^2 \norm{u_1 - u_2}^2$, and $\Delta_t m = m_{t+1} - m_t$.
This discrete-time evolution can be solved recursively.
A nice formula can be derived for the continuous-time version, i.e. for the flow \eqref{eq:GF} instead of the descent \eqref{eq:GD}, where
\[
    (1 + \exp(m))\diff m = c_x\diff t.
\]
In particular when $w$ is initialized at zero,
\[
    m_t + \exp(m_t) = c_x t.
\]
This equation is inverted with the product logarithm, giving an exact expression for the margin evolution, proven in Appendix \ref{proof:bin-ortho}.

\begin{theorem}[Binary orthogonal]
    \label{thm:ortho-bin}
    Let $M=2$, and the input embeddings be orthogonal.
    The dynamics \eqref{eq:GF}, \eqref{eq:GD}, \eqref{eq:SGF} and \eqref{eq:SGD} lead to
    \begin{equation}
        W_t = \sum_x m_t(x) \frac{u_1 - u_2}{\norm{u_1 -u_2}^2} \otimes \frac{e_x}{\norm{e_x}^2} + \Pi_\perp W_0,
    \end{equation}
	where $\Pi_\perp$ is the projection on the orthogonal of the span of the gradient updates.
	For gradient flow \eqref{eq:GF}, there exists a $t_0\in\R$ that depends on initial condition, e.g. $t_0 = -1/c_x$ when $W_0 = 0$, such that when $t \geq t_0$, the exact evolution is given by,
    \begin{equation}
        \label{eq:dyn-bin}
        m_t(x) = \log(c_x (t-t_0)) - h(c_x (t-t_0)),
    \end{equation}
    where $c_x = p(x) \norm{e_x}^2\norm{u_1-u_{2}}^2$ and $h$ is a function such that $0 \leq h(x) \leq 2 \log(x) / x$ for all $x \geq 1$.
	Similarly, for gradient descent \eqref{eq:GD} with a learning rate $\eta$,
    \[
        m_t(x) \geq \log(\eta c_x (t-t_0)).
    \]
	In particular, when $W_0 = 0$, it leads to the following bound on the loss,
	\[
        \cL(W_t) \leq \frac{1}{t\eta} \sum_x \frac{p(x)}{c_x}.
	\]
\end{theorem}

The setting of Theorem \ref{thm:ortho-bin} is a special case of logistic regression with linearly separable data where all margins grow logarithmically without inhibiting each other. 
The loss monotonically decays in $O(1 / \eta t)$ which corresponds to the rate of \citet{wu2023implicit}, with the addition of the explicit dependence on the learning rate.

\paragraph{Multi-class.}
We now attack the multi-class case. 
Let
\[
    w_i = w_{xi} = u_i^\top W e_x,
\]
and denote the partition function $A(w) = \sum \exp(w_i)$ with $w = (w_i)$.
Let us focus for now on gradient flow for simplicity. When the $(u_i)$ are orthogonal, the dynamics \eqref{eq:interaction} becomes
\begin{align*}
	&\frac{A(w)}{A(w) - \exp(w_1)} \diff w_1 = p(x)\norm{e_x}^2\norm{u_1}^2 \diff t,
	\\& A(w) \exp(-w_i) \diff w_i = -p(x)\norm{e_x}^2\norm{u_i}^2\diff t,
\end{align*}
for $i \neq 1$. 
In the multi-class case, the evolution of the margins $m_i = w_1 - w_i$ do not directly decouple from each other as in the binary case. 

One can combine these differential equations to find many invariants of this system of interacting particles.
In particular, $\exp(-w_i) - \exp(-w_j)$ stays constant over time.\footnote{This generates all the invariants of the dynamics but one, the remaining one is a consequence of $\diff W \in \Span\brace{u_i - u_j}_{ij} \otimes R^d$.}
Some algebraic manipulations implies the following evolution of the tightest margin (corresponding to $\argmax_{i\neq 1} p_W(i|x)$)
\[
    (c_i + \exp(m_i)) \diff m_i = p(x)\norm{e_x}^2(c_i + 1) \diff t.
\]
for some bounded function $c_i$ that depends on initial conditions.
This leads to the same logarithm convergence as in the binary case.
For simplicity, we only report the asymptotic behavior in Theorem \ref{thm:multi}, which we prove in Appendix~\ref{proof:ortho}.

\begin{theorem}[Multi-class orthogonal]
    \label{thm:multi}
    Assume that the input and output embeddings are orthonormal.
    For any initialization, the gradient flow dynamics \eqref{eq:GF} converges as
    \begin{equation*}
        \label{eq:lim-bin}
        \lim_{t\to\infty}\frac{W_t}{\log(t)}
        \propto \sum_x \Pi(u_{f^*(x)}) \otimes e_x,
    \end{equation*}
    where $\Pi$ is the projection on the span of the $(u_i - u_j)$.
\end{theorem}

For gradient descent, one can express the updates for the margins similarly
\[
    A(w)\Delta_t m_i = \eta_t(A(w) - e^{w_1} + e^{w_i}) p(x) \| e_x \|^2 > 0.
\]
Hence, the margins only increase during training, and the larger the learning rate, the faster the evolution.
Indeed, one gradient step is enough to learn all the associations $x \to f^*(x)$, i.e. $f_{W_1}(x) = f^*(x)$ for any initialization. 
Continuing the training will continue to increase the margins, ultimately ensuring the convergence of $W_t$ to the max-margin solution of the classification problem, as characterized by Theorem \ref{thm:multi}, making the final classifier robust to embedding displacements \citep{Cortes1995}.

To conclude, when the embeddings are orthogonal, the memories do not interfere much, and one can learn all the associations with one giant gradient step.
This case still presents several behaviors of interest.
First of all, Equation~\eqref{eq:dyn-bin} shows that the association $x \to y$ is learned faster when $x$ is frequent, i.e. $p(x)$ is large.
Indeed, early in training, one can envision $W \simeq \sum_x p(x) u_y \otimes e_x$.
However, later in training, the training dynamics will start saturating in the direction $u_y \otimes e_x$ for the frequent tokens, allowing the less frequent ones to catch up.
The catch-up is facilitated by large learning rates. 
Ultimately, as shown by Equation~\eqref{eq:lim-bin}, the final $W$ does not depend on the token frequencies (see~\citealp{byrd2019effect} for related observations).
In other terms, if the model has enough capacity to learn all the data (in our case, orthogonality implies $d \geq N$), then at the end of the training, it allocates equal capacity to every token even though some tokens are much rarer.
Nonetheless, curating data to make them less redundant can make learning more efficient.

\subsection{Particles Interfering}

Let us now consider the case where $N\leq d$, but where memories interfere between them.
We first notice that in the case when the input embeddings are orthogonal, correlated output embeddings introduce limited competition, and this case can largely be understood as a simplified version of the interaction between input embeddings.

\begin{figure}[t]
    \centering
    \vspace{-.35em}
    \includegraphics{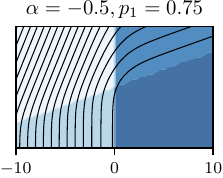}
    \hspace{.5em}
    \includegraphics{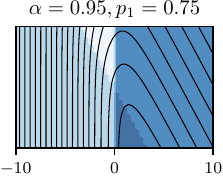}
    \vspace{-.5em}
    \caption{Level lines of $\cL(W)$ for $N=d=2$ as a function of $\gamma_i(W):= (u_2-u_1)^\top Wf_i$ where $(f_i)$ is a basis of $\R^2$. 
    Token embeddings have correlation $\alpha$ \eqref{eq:correlation-binary}.
    We equally plot the value of $\cL_{01}(W)$, dark blue meaning perfect accuracy, and white meaning null accuracy.}
    \label{fig:level}
    \vspace{-1em}
\end{figure}

Let us analyze the simple but instructive case $N = 2$ of two input tokens with $f^*(x) = x$, and
\begin{equation}
    \label{eq:correlation-binary}    
    \alpha_{ij} = \scap{e_i}{e_j} = \ind{i=j} + \alpha \ind{i\neq j},\qquad \alpha \in [-1,1].
\end{equation}
In other terms, the input embeddings are normalized and are $\alpha$-correlated.
Two margins are at play:
\[
    m_i = w_{ii} - w_{ij} = (u_i - u_j)^\top W e_i, \quad \{i,j\}=\{1,2\}.
\]
The interacting system \eqref{eq:interaction} becomes,
\begin{equation}
    \label{eq:two-descent}    
    \Delta m_i = c \eta_t \paren{\frac{p_i}{1 + \exp(m_i)} - \frac{\alpha p_j}{1 + \exp(m_j)}},
\end{equation}
where $c= \| u_1 - u_2 \|^2$ and we denote $p_i = p(i)$ for readability.
In the gradient dynamics, $x=1$ pushes $W$ in the direction $(u_1 - u_2)\otimes e_1$, which, when $\alpha \leq 0$, is positively correlated with the direction $(u_2 - u_1)\otimes e_2$ promoted by $x=2$.
As can be seen in Equation~\eqref{eq:two-descent}, when $\alpha \leq 0$, both margins increase during training, there is no competition between the memories, and a single gradient step is enough to reach perfect accuracy.
\begin{figure}[t]
    \vspace{-.35em}
    \centering
    \includegraphics{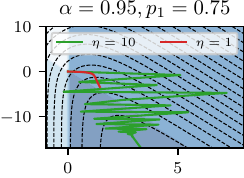}
    \includegraphics{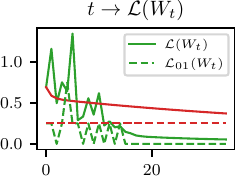}\\
    \vspace{-.5em}
    \caption{{\em Loss spikes.} Trajectories of $W_t$ in the setting of Figure \ref{fig:level} for two learning rates $\eta$, $\eta=10$ in green, $\eta=1$ in red, and their traces in term of losses as a function of the number of epochs, here $t\in[35]$.}
    \label{fig:translate}
    \vspace{-1em}
\end{figure}
To solve Equation \eqref{eq:two-descent}, let us introduce the orthogonal family
\[
    f_1 = e_1 + e_2, \qquad
    f_2 = e_1 - e_2,
\]
and project the dynamics on those directions with
\begin{equation}
    \gamma_i = \frac{1}{2}(u_1 - u_2)^\top W f_i.
\end{equation}
The evolution of the $\gamma_i$ is governed by
\begin{align}
    \nonumber
    & \frac{\diff \gamma_1}{\diff ct} 
    =  \frac{(1+\alpha)p_1}{1 + \exp\paren{\gamma_2 + \gamma_1}} - \frac{(1+\alpha)p_2}{1 + \exp\paren{\gamma_2 - \gamma_1}}
   \\ &\frac{\diff \gamma_2}{\diff ct} 
    = \frac{(1-\alpha)p_1}{1 + \exp\paren{\gamma_2 + \gamma_1}} + \frac{(1-\alpha)p_2}{1 + \exp\paren{\gamma_2 - \gamma_1}},
    \label{eq:two-input}
\end{align}
From the second differential equation, we see that $\gamma_2$ always increases during the dynamics.
The growth of $\gamma_2$ will slow down the growth of $\gamma_1$. 
These together imply that $W$ grows logarithmically in one direction ($f_2$, which turns out to be the max-margin direction) and stays bounded in the orthogonal direction, which we prove in Appendix~\ref{app:two-tokens-interference} and is the object of the following theorem.

\begin{theorem}[Two particles interacting] 
    \label{thm:two-flow}
    Let $N=2$ with $f^*(x) = x$.
    Assume without restriction that $p_1 \geq p_2$.
    When Equation \eqref{eq:correlation-binary} holds, if $W$ is initialized at zero, i.e. $W_0=0$, for gradient flow,
    \begin{align*}
        &\gamma_2(t) = \log(c_t t +1) + O\paren{\frac{\log(c_2 t +1)}{c_2 t +1}}, \\
        &\gamma_1(t) = \frac{1}{2} \log\parend[\big]{p_1/p_2} + O(1/t),
    \end{align*}
    where $2p_2 \leq c_t / c_0 \leq 8 p_1^3 / p_2^2$ with $c_0=(1-\alpha) c$.
    Similarly for gradient descent, with any step-size~$\eta \geq 0$,
    \begin{align*}
        &\gamma_2(t) \geq \log(\eta p_2 c_0 t +1) + O\paren{\frac{\log(t)}{t}}, \\
        &\abs{\gamma_1(t) - \frac{1}{2}\log(p_1 / p_2)} \leq \eta(1 - \alpha) p_1 + \frac{p_1}{2p_2} + O(1/t).
    \end{align*}
\end{theorem}

These results are consistent with~\citet{wu2023implicit,wu2024large}, although our focus is to obtain a fine-grained dependence on the quantities relevant to our setting ($\alpha, p_1, p_2, \eta$).
For any learning rate~$\eta$, when~$t$ grows large, both margins eventually become positive (since they are proportional to~$\gamma_2 \pm \gamma_1$ with~$\gamma_1$ bounded), leading to perfect accuracy of our model.

In the dynamics analyzed so far, we observe a stationary regime where $W_t \simeq \log(t) W_\infty$.
However, transitory regimes can hide under the big-$O$ in Theorem \ref{thm:two-flow} --we characterize the big-O precisely in the appendix.
When considering discrete-time dynamics such as gradient descent \eqref{eq:GD}, or stochastic dynamics, i.e., \eqref{eq:SGF} or \eqref{eq:SGD}, those transitory regimes can showcase weight oscillations and loss spikes.
For example, when $N=2$ and there is strong association imbalanced and correlation, viz. $\alpha p(1) \gg p(2)$, the dynamics at the beginning of training can be approximated by
\begin{align}
    \nonumber
    \Delta m_1 = \frac{\eta_t p(1)}{1 + \exp(m_1)}, \quad
    \Delta m_2 = \frac{- \eta_t \alpha p(1)}{1 + \exp(m_1)}.
\end{align}
Hence, in terms of the association stored in $W$, when the learning rate is large, the token $x=1$ will ``erase'' the token $x=2$.
Since $p_W(f^*(x)|x=2)$ approaching to zero implies that $\ell(W;x,f^*(x))$ goes to infinity, this can lead to arbitrarily big loss spikes, as captured by Proposition \ref{thm:spike}, proved in Appendix \ref{proof:spike}.
However, later in training, $p_W(2)$ catches up and $W$  ultimately aligns in the max-margin direction, while $m_1 - m_2$ remains bounded.

\begin{proposition}[Loss spikes]
    \label{thm:spike}
    Let $N=2$ with $f^*(x) = x$.
    Assume that Equation \eqref{eq:correlation-binary} holds, and $\alpha p_1 - p_2>0$. 
    From a null initialization $W_0=0$, one gradient update \eqref{eq:GD} with learning rate $\eta$ leads to 
    \begin{equation}
        \label{eq:spike}
        \cL(W_1) \geq \eta(\alpha p_1 - p_2) p_2,
    \end{equation}
    which can be arbitrarily large.
\end{proposition}

\begin{figure}[t]
    \vspace{-.35em}
    \centering
    \includegraphics{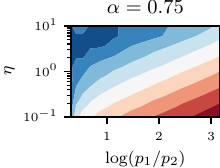}
    \hspace{.5em}
    \includegraphics{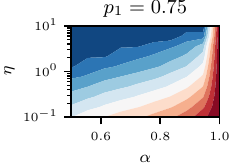}
    \caption{
        Level lines of the (logarithm of the) number of steps needed to reach perfect accuracy in the setting of Theorem \ref{thm:two-flow}, as a function of the learning rates $\eta$, the interaction parameter $\alpha$, and the class imbalance $\log(p_1 / p_2)$.
        Red means more steps to reach perfect accuracy.
    }
    \label{fig:bin-int}
    \vspace{-1.5em}
\end{figure}

To conclude, for overparameterized models, the dynamics is initially governed by memory interactions, before settling in a stationary regime similar to the orthogonal case described in Theorem~\ref{thm:multi}. 
The oscillatory regime is due to the competition between two groups of tokens where increasing the margins of the high-frequency tokens causes a decrease in the margins of the others, similar to the opposing signals in \citet{rosenfeld2023outliers}.
The settling down of the dynamics can be understood intuitively.
Since the max-margin will grow, all the partition function $A(We_x)$ of the $p_W(y|x)$ will grow, which will slow down the dynamics. Hence the oscillation will fade, and dynamics will enter the stationary logarithmic regime.
In the stationary regime, bigger learning rates act as a speed-up of time, ensuring faster convergence.
From a learning efficiency point of view, there is a trade-off between large learning rates implying longer oscillatory transitory regimes, and small learning rates implying slow speed of the dynamics. 
We illustrate this trade-off on Figure~\ref{fig:bin-int}.
We observe that class imbalance, and interference makes the problem harder, and that large learning rates is beneficial, although very large learning rates can be detrimental (top left of the left plot).

\subsection{Graphical Understanding}

Now that we have a good understanding of the mechanisms at play, we can verify these phenomena more generally through simulations.
Let us first leverage the previous derivations to explain how to read measures of performance that can be obtained from experiments.
When $d=M=2$ and any value of $N\geq 2$, the problem reduces to a two-dimensional ones with
\begin{equation}
    \label{eq:fig-cor}    
    \gamma_i = (u_2 - u_1)^\top W f_i, \quad (f_i)_k = \ind{i=k}.
\end{equation}
In the resulting two-dimensional space, we can plot the level lines of the loss function, the level lines of its Hessian eigenvalues, as well as the trajectories followed by $W_t$ for different optimizers.

\begin{figure*}[t]
    \vspace{-.35em}
    \centering
    \includegraphics{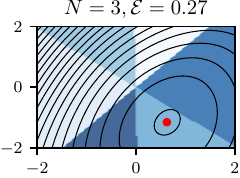}
    \includegraphics{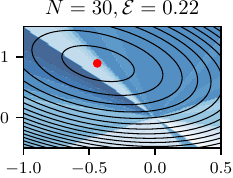}
    \includegraphics{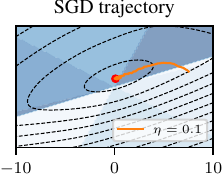}
    \includegraphics{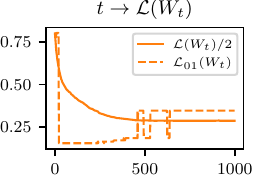}\\
    \vspace{-.5em}
    \caption{{\em Forgetting.} Similar plots as in Figures \ref{fig:level} and \ref{fig:translate}, yet in the limited capacity case $d < N$.
    In those situations, competition between the memories can lead to sub-optimal minimizer of $\cL$, which we illustrate with SGD on the bottom plots.
    The sub-optimality is reflected in the excess of risk $\cE = \cL_{01}(\argmin_W \cL(W)) - \min_W\cL_{01}(W)$. 
    }
    \label{fig:forgetting}
\end{figure*}

\begin{figure*}[t]
    \centering
    \includegraphics{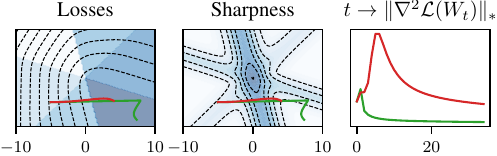}
    \includegraphics{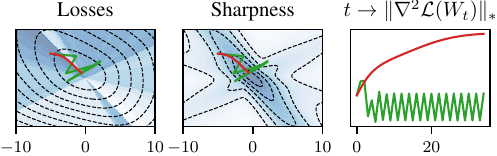}
    \vspace{-.5em}
    \caption{{\em Sharpness profile.} Gradient descent trajectories in the setting of Figures \ref{fig:translate} and \ref{fig:forgetting} with learning rates $\eta=10$ (green) and $\eta=1$ (red). We plot the level lines of the sharpness, i.e. the operator norm of $\nabla^2 \cL(W)$, as well as the trace of the trajectories in terms of sharpness. The left plots are in the overparameterized regime, the right ones in the underparameterized one.}
    \label{fig:sharpness}
    \vspace{-.5em}
\end{figure*}

Figure \ref{fig:level} shows the level lines of the loss for $p(1) = \sfrac{3}{4}$ and $\alpha \in \brace{-\sfrac{1}{2}, 0.95}$.
The deterministic gradient trajectories can be deduced from this picture: they are always orthogonal to level lines, and their speed is proportional to the number of lines crossed locally.
The fact that there is not much level line in the region $\brace{W\vert \cL_{01}(W) = 0}$ is due to the logarithmic convergence illustrated by Theorem \ref{thm:multi}.
The right of Figure \ref{fig:level} shows that, although gradients are always positively correlated with the max-margin direction (formally $\diff \gamma_2 \geq 0$ in \eqref{eq:two-input}), they can point in directions that do not lead to perfect accuracy.
Indeed, Figure \ref{fig:translate} illustrates how large learning rates are likely to result in spikes of both the loss and the accuracy.
This latter figure shows the trajectories of $W_t$ for two different learning rates, and the trace of these trajectories in the training loss and accuracy plots which are usually monitored by practitioners training neural networks.

\section{Numerical Analysis}

This section complements previous derivations with numerical analysis.
It discusses underparameterized regimes, large versions of model \eqref{eq:model}, as well as more complicated ones.

\subsection{Limited capacity}
Let us start the numerical analysis with the case where $N > d$.
In those cases, one can not necessarily store all associations in memory, and the model has to favor some of them.
It was shown in \citet{cabannes2023scaling} that the ideal $W$ can usually store about $d$ memories similarly to Hopfield network scalings.
However, this ideal $W$ is not always the one minimizing the cross-entropy loss.

We plot our problem in the case $M=d=2$ thanks to the statistics $\gamma_i$ of \eqref{eq:fig-cor}.
Figure \ref{fig:forgetting} reveals a striking fact: the cross-entropy loss is not calibrated for our model, i.e., minimizing $\cL(W)$ does not always minimize $\cL_{01}(W)$.
Indeed, even in the case $N=3$, one can find examples where competition between the memories leads the minimizer of $\cL$ to ``forget'' the most frequent association.
When $N$ becomes large in front of $d$, these cases become the norm.
On these landscapes, one can come up with examples of catastrophic forgetting, where the dynamics is first dominated by frequent tokens that are well memorized until rare classes come into play, perturbing the minimizer of $\cL$, ultimately leading to convergence to a sub-optimal place.
We illustrate it on the right of Figure \ref{fig:forgetting}.

To further illustrate the differences between the dynamics in over- and under-parameterized regimes, Figure~\ref{fig:sharpness} illustrates the sharpness, as defined by the operator norm of the Hessian of $\cL(W)$ along two descent trajectories.
We compute the Hessian in closed-form to show its level lines, illustrating that the sharpness of the logistic loss is mainly high for small values of the norm of $W$.
We observe three types of behaviors.
In the separable case, e.g. when $d \geq N$, the transitory regime goes through relatively sharp regions, before the stationary regime where the sharpness decreases until reaching zero at infinity.
In the non-separable case, with is typical when $d < N$, either the learning rate is small enough and we converge to the minimum of $\cL$ presenting a sharpness $H_*$ greater than $2/\eta$, or the learning rate is greater than $2/H_*$ and we oscillate around the minimizer of~$\cL$.

\subsection{Larger dimension}

\begin{figure}[t]
    \vspace{-.35em}
    \centering
    \includegraphics{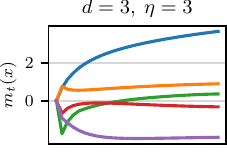}
    \hspace{.5em}
    \includegraphics{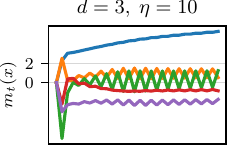} \\
    \vspace{.5em}
    \includegraphics{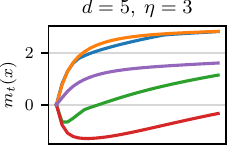}
    \hspace{.5em}
    \includegraphics{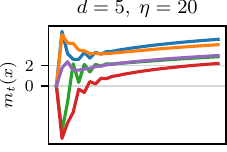} \\
    \vspace{.5em}
    \includegraphics{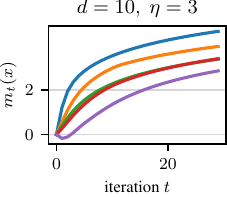}
    \hspace{.5em}
    \includegraphics{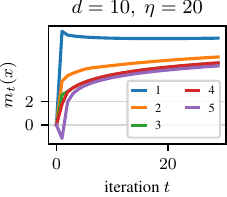} \\
    \caption{Margins~$m_t(x)$ for~$N = 5$ tokens, with varying dimensions~$d$ and learning rates~$\eta$. The embeddings were sampled uniformly at random on the sphere.
    Large learning rates learn faster, although they lead to more oscillation, especially in low dimension.
    When $d < N$, the model does not have enough capacity to learn all the associations, and it favors the most frequent ones.
    }
    \label{fig:margins_N5}
    \vspace{-1.5em}
\end{figure}

When the dimension~$d$ is larger, although we can not plot the weight-space, we can plot the evolution of certain statistics, such as the margin, along descent trajectories.
In Figure~\ref{fig:margins_N5}, we consider a setup with~$N = M = 5$,~$f^*(x) = x$, and~$p(x) \propto 1/x$, in different dimensions (with random embeddings). 
We show the evolution of the margins
\begin{equation}
    m_t(x) = \langle u_{f^*(x)}, W e_x \rangle - \max_{j \ne x} \langle u_j, W e_x \rangle.
\end{equation}
Perfect accuracy is achieved when~$m_t(x) > 0$ for all~$x$.
We see the faster increase of margins for more frequent tokens, faster convergence with large step-size~$\eta$, at the cost of oscillations, and benefits of larger~$d$. 
The latter are likely due to less interference thanks to more orthogonality between random embeddings in higher dimension.

\subsection{Simplified Transformer model}
\label{sub:transformer}

\begin{figure}[t]
    \vspace{-.35em}
    \centering
    \includegraphics{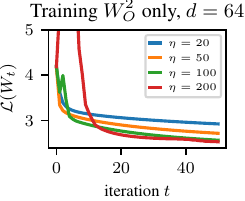}
    \includegraphics{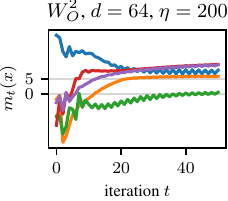} \\
    \vspace{2em}
    \includegraphics{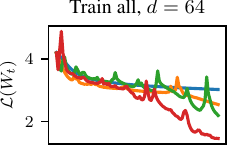}
    \hspace{.5em}
    \includegraphics{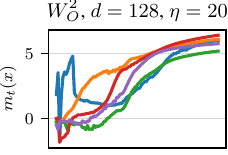} \\
    \vspace{.5em}
    \includegraphics{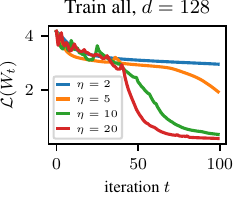}
    \hspace{.5em}
    \includegraphics{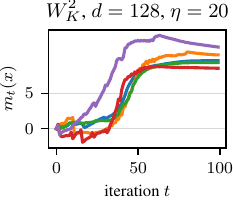}
    \caption{Full-batch training of selected transformer layers on the bigram task.
    (top) Loss and margins when training~$W_O^2$ alone.
    (bottom) Loss and margins when training the three layers ($W_O^2$, $W_K^2$, and~$W_K^1$) sufficient for the task, for two widths~$d$.
    Training losses are shown for different step-sizes~$\eta$, and margins are shown for~5 different tokens.}
    \label{fig:transformer}
    \vspace{-1.5em}
\end{figure}

Finally, we empirically study training dynamics on a more complex model involving multiple associative memory mechanisms like the ones above. 
In particular, we consider a simplified two-layer Transformer architecture trained on an in-context learning task (described in Appendix~\ref{sub:appx_transformer}) that requires copying a bigram from the context depending on the current token. 
A two-layer attention-only transformer can solve this by implementing an ``induction head'' mechanism~\citep{elhage2021mathematical,olsson2022context}, and~\citet{bietti2023birth} show that this can be achieved by training only three matrices~$W_O^2$, $W_K^2$, and~$W_K^1$. These were found to behave as associative memories with appropriate embeddings, and we may thus empirically assess their margins.

We consider full-batch gradient descent on a dataset of 16\,384 sequences of length 256 generated from the model described above with~$N=64$ tokens.
The top of Figure~\ref{fig:transformer} shows the training loss and margins when only training~$W_O^2$, which can learn an appropriate associative memory by itself.
The objective for this problem is convex and similar to the ones considered in this paper, up to some noise in the input embeddings due to attention.
We see that margins tend to increase during training, and that large learning rates lead to faster optimization of the loss, at the cost of some spikes in the loss, and oscillations in the margins, which are due to correlations between embeddings.

The bottom of Figure~\ref{fig:transformer} shows loss curves and margin evolution when training all three matrices. 
Here we see more frequent spikes in the loss for large learning rates, yet their gains are much more significant later in training, with small final losses that suggest the induction head mechanism is learned. The increasing margins confirm that the desired associative behaviors have indeed been recovered. 
Compared to the top of Figure~\ref{fig:transformer}, the margins for~$W_O^2$ display more significant oscillations initially, likely due to additional interactions across different parameter matrices. 
In later iterations, when the attention heads are in place and inputs to~$W_O^2$ are less noisy, the margins increase together to large values, leading to a similar learning speed on all memories. 
This uniform convergence behavior was facilitated by the relatively low output tokens imbalance considered in our tasks where the ``copied'' tokens were sampled uniformly.
Finally, we see the effect of larger embedding dimensions~$d$, accelerating the convergence thanks to more orthogonality.

\section{Discussion}

In this paper, we studied the gradient dynamics of associative memory models trained with cross-entropy loss, by viewing memory associations as interacting particles.
This leads to new insights on the role of the data distribution and correlated embeddings on convergence speed as well as training ``instabilities'' in large learning rate regimes, such as oscillations and loss~spikes.
We also showed that some of these insights may transfer to some more realistic scenarios such as training small Transformers.
Nonetheless, our simple model is only a first step, and there are many additional factors at play in larger models, which may lead to different behaviors and instabilities. This includes factorized parameterizations, normalization layers, adaptive optimizers, noisy data, and interactions between different layers which may change at different timescales.
Studying the impact of these on training dynamics could unlock new improvements to the practice and reliability of training large models.


\subsection*{Acknowledgements}
The authors would like to thank L\'eon Bottou and Herv\'e Jegou for fruitful discussions that led to this line of work.

\bibliography{reference}
\bibliographystyle{template/icml2024}

\appendix
\onecolumn
\section{Gradient derivations}
\label{app:derivative}

In the following, to be consistent with pytorch convention, we redefine the model as
\[
    f_W(x, y) = \scap{e_x}{Wu_y}
\]
Recall that the loss can be understood intuitively as the negative log-likelihood
\[
    \cL(W) = -\E_{X,Y}[\log p_W(Y\,|\, X)], \qquad\text{where}\qquad
    p_W(y \,|\, x) = \frac{\exp(\scap{e_x}{W u_y})}{\sum_z \exp(\scap{e_x}{W u_z})},
\]
of the probability $p_W$ whose conditional distributions are parameterized as a soft-max over the score $\scap{e_x}{W u_y}$.
In information theory, this negative log-likelihood is known as the cross-entropy of the model probability $p_W$ relative to the real data distribution $p$.

To analyze the training dynamics, we will monitor quantities related to the gradient and the Hessian of the loss function.
The gradient of the loss is easy to compute with simple derivation rules.
\begin{align*}
    \nabla \ell(W; x, y) 
    &= \nabla \log \left( \sum_{z} \exp \langle e_x, W u_z \rangle \right) - \nabla \langle e_x, W u_y \rangle
    \\&= \sum_{z} \frac{\exp \langle e_x, W u_z \rangle}{\sum_{z'} \exp \langle e_x, W u_{z'} \rangle} e_x u_z^\top  - e_x u_y^\top.
\end{align*}
This gradient can be understood with the probabilistic perspective on the loss as
\begin{equation}
    \nabla \ell(W; x, y) 
    = \sum_{z} p_W(z\,\vert\, x) e_x u_z^\top - e_x u_y^\top.
\end{equation}

\subsubsection*{Hessian computation}
Notice that when the gradient can be written as
\[
    \nabla \ell(\theta) = g(\theta) \cdot a = (\partial_i \ell(\theta))_{i}, \qquad\text{with}\qquad g(\theta) \in \R,\quad a \in \R^d,
\]
the Hessian follows as
\[
    \nabla^2 \ell(\theta) = (\partial_{ij}\ell(\theta)) = (\partial_i\partial_j \ell(\theta)) = (\partial_i g(\theta)\, a_j) = (\nabla g(\theta))\,a^\top.
\]
In our case, we want to use the Euclidean structure on the matrix space, which leads to
\begin{equation}
    \label{eq:hess_der1}
    \nabla^2 \ell(W; x, y) = \sum_{z} \nabla p_W(z\,\vert\, x) (e_x \otimes u_z)^\top.
\end{equation}
To compute $\nabla p_W(z\,\vert\, x)$, notice that we could equally have expressed the loss gradient as
\[
    \nabla \ell(W; x, y) = -\nabla \log p_W(y\,\vert\,x) = -\frac{\nabla p_W(y\,\vert\,x)}{p_W(y\,\vert\, x)},
\]
from which we deduce that
\begin{align}
    \nabla p_W(z\,\vert\,x) 
    &= - p_W(z\,\vert\, x) \nabla \ell(W; x, z)
    \notag
    \\&= p_W(z\,\vert\, x)\left(e_x \otimes u_z - \sum_{z'} p_W(z'\,\vert\, x) e_x \otimes u_{z'} \right).
    \label{eq:hess_der2}
\end{align}
Plugging this into Equation \eqref{eq:hess_der1}, we will have to deal with quantities such as\footnote{
    Recall that a tensor $M = a_1 \otimes a_2 \ldots \otimes a_p \in (\R^d)^{\otimes p} $ can be understood as a $p$-dimensional matrix ${\tt M}$.
    If $(e_i)$ denotes the basis of $\R^d$, then $e_{j_1} \otimes e_{j_2} \ldots \otimes e_{j_p}$ is a basis of the tensor space, and $M$ assimilates to ${\tt M}$ such that ${\tt M[j_1, j_2, \ldots, j_p]} = \prod_{i\in[p]}\scap{a_i}{e_{j_i}}$.
}
\[
    (e_x \otimes u_y) (e_x \otimes u_z)^\top = e_x \otimes u_y \otimes e_x \otimes u_z.
\]
The last operation can be understood from the fact that, when $f_i$ is the canonical basis of $\R^d$, and $a$ and $b$ are in $\R^d$, we have the matrix identification
\[
    (a b^\top)_{i,j} = \langle a, f_i \rangle \langle b, f_j \rangle  
\]
In our case, using $ij$ as the matrix indexation for $\R^{d\times d}$ and $f_i f_j^\top$ as the canonical basis of $\R^{d\times d}$,
\begin{align*}
    \left((e_x \otimes u_y)(e_x\otimes u_z)^\top\right)_{ij, kl}
    &=  \langle e_x \otimes u_y, f_i \otimes f_j \rangle \langle e_x\otimes u_z, f_k\otimes f_l \rangle  
    \\&= (f_i^\top e_x)(f_j^\top u_y)(f_k^\top e_x)(f_i^\top u_z)
    \\&= \langle f_i\otimes f_j\otimes f_k\otimes f_l\otimes, e_x\otimes u_y\otimes e_x\otimes u_z\rangle.
\end{align*}

Using Equations \eqref{eq:hess_der1} and \eqref{eq:hess_der2}, we deduce
\begin{equation}
    \label{eq:hess}
    \nabla^2 \ell(W; x, y) 
    = \sum_{z,z'} p_W(z\,\vert\, x) (\delta_{z,z'} - p_W(z'\,\vert\, x)) e_x\otimes u_z\otimes e_x\otimes u_{z'}.
\end{equation}
We implemented this formula vectorially in our code, and checked its correctness based on automatic differentiation libraries.

\section{Dynamics without interference}
\label{app:dyn-ortho}

To study the dynamic, using Equation \eqref{eq:grad}, with the notation of the main text, i.e. $f_W(x,y) = \scap{u_y}{W e_x}$, we have
\[
    \nabla \ell(W; x, y) e_i
    = \sum_{z} p_W(z\,\vert\, x) (u_{z} - u_y) e_x^\top e_i.
\]
In particular, when the $(e_i)$ are orthogonal, summing over $x$ leads to
\begin{equation}
    \label{eq:ortho-dyn-tmp1}    
    \nabla\cL(W)e_x = p(x) \norm{e_x}^2 \sum_{z} p_W(z\,\vert\, x) (u_{z} - u_{f^*(x)}).
\end{equation}

\subsection{Binary classification - Proof of Theorem~\ref{thm:ortho-bin}}
\label{proof:bin-ortho}

Let us consider the binary case where $y\in\brace{1,2}$.
Assume that $f^*(x) = 1$, Equation \eqref{eq:ortho-dyn-tmp1} simplifies as
\[
    \nabla\cL(W)e_x = p(x) \norm{e_x}^2 p_W(2\,\vert\, x) (u_2 - u_1).
\]
We can project it on the line where it evolves, reducing this equation to a scalar evolution
\begin{align*}
    (u_1 - u_2)^\top\nabla\cL(W)e_x 
    &= -p(x) \norm{e_x}^2 p_W(2\,\vert\, x) \norm{u_2 - u_1}^2
    \\&= - p(x) \norm{e_x}^2 \norm{u_2 - u_1}^2 \frac{\exp(u_2^\top W e_x)}{\exp(u_1^\top W e_x) + \exp(u_2^\top W e_x)}
    \\&= - p(x) \norm{e_x}^2 \norm{u_2 - u_1}^2 \frac{1}{\exp((u_1 - u_2)^\top W e_x) + 1}.
\end{align*}
Let us consider the evolution equation, for some learning rate scheduling $(\eta_t)_{t \geq 1}$
\[
    W_{t+1} = W_t - \eta_t \nabla \cL(W_t).
\]
This leads to
\[
    (u_1-u_2)^\top (W_{t+1} - W_t)e_x = \eta_t p(x)\norm{e_x}^2 \norm{u_2 - u_1}^2 \frac{1}{\exp((u_1 - u_2)^\top W_t e_x) + 1}.
\]
Let us set
\begin{equation*}
    m_t = (u_1 - u_2)^\top W_t e_x, 
    \qquad c = p(x) \norm{e_x}^2 \norm{u_2 - u_1}^2.
\end{equation*}
The evolution equation becomes
\begin{equation}\label{eq:binary-update}
    (\exp(m_t) + 1) (m_{t+1} - m_t) = \eta_t c.
\end{equation}

\subsubsection{Gradient flow.}
Since the updates are in the span of $(u_1 - u_2) \otimes e_x$, we have 
\[
    W_t = \sum_x \alpha_t(x) (u_1 - u_2) \otimes e_x + \Pi_\perp W_0,
\]
where $\Pi_\perp$ is the projection on the orthogonal of the gradient updates, and $\alpha_t(x)$ can be inferred from the margin
\[
    m_t(x) = (u_1-u_2)^T W_t e_x = \alpha_t(x) \| u_1 - u_2 \|^2 \| e_x \|^2,
\]
In the following, we will solve the ODE for each margin using the product logarithm.
The gradient flow limit of the previous derivation in Eq.~\ref{eq:binary-update} leads to 
\[
   (\exp(m) + 1) \diff m = c\diff t.
\]
Integrating this differential equation gives
\[
    \exp(m_t) + m_t = c t + \exp(m_0) + m_0.
\]
For $x, y\in\R$, we can solve
\[
    \exp(x) + x = y
    \quad\Leftrightarrow\quad
    (y - x)\exp(y-x) = \exp(y)
    \quad\Leftrightarrow\quad
    x = y - \cW_0(e^y),
\]
where $\cW_0$ is the product logarithm.
This allows us to solve the previous equation in closed-form
\begin{equation}
    m_t = c t + e^{m_0} + m_0 - \cW_0(e^{c t} e^{\exp(m_0) + m_0}). 
\end{equation}
We can simplify this profile using the following asymptotic development of the product logarithm \cite{hoorfar2008inequalities} when $x \geq e$,
\[
    \cW_0(x) = \log(x) - \log \log(x) + \frac{\tilde h(\log(x))\log\log(x)}{\log(x)}, \qquad\text{with}\qquad
    \tilde h(x) \in [\sfrac{1}{2}, 2].
\]
We deduce that as soon as $ct + \exp(m_0) + m_0 \geq 0$,
\[
    m_t = \log(ct + \exp(m_0) + m_0) - \frac{\tilde{h}(ct + \exp(m_0) + m_0)\log(ct + \exp(m_0) + m_0)}{ct + \exp(m_0) + m_0},
\]
since
\[
  \cW_0(e^{c t} e^{\exp(m_0) + m_0}) = ct + \exp(m_0) + m_0 - \log(ct + \exp(m_0) + m_0) + \frac{\tilde{h}(\cdots)\log(ct + \exp(m_0) + m_0)}{ct + \exp(m_0) + m_0}.
\]
The first part of theorem is found with 
\[
  t_0 = -\frac{\exp(m_0) + m_0}{c},
\]
and with the substitution of $\tilde h$ by $h(x) = \tilde{h}(x)\log(x) / x$.

\subsubsection{Gradient descent}
In the case of gradient descent, we will work with the discrete update equation 
\[
    m_{t+1} - m_t = \frac{c \eta_t}{\exp(m_t) + 1}.
\]
Since we expect a logarithmic growth of the margins, we exponentiate this equation and rearrange terms
\[
    \exp(m_{t+1}) - \exp(m_t) = \exp(m_t) \left( \exp \bigl(\frac{c \eta_t}{\exp(m_t) + 1}\bigr) - 1\right). 
\]
Notably, when we initialize the weights at zero, all margin updates are positive in this case with no interferences. 
This implies that $m_t \geq 0$ at all times. 
\[
    \exp(m_{t+1}) - \exp(m_t) \geq c \eta_t \frac{ \exp(m_t) }{\exp(m_t) + 1} \geq c \eta_t \frac{1}{2}
\]
where we used $e^x \geq x+1$ and $e^x/(e^x+1)\geq 1/2$ for $x \geq 0$. 
Telescoping this summation, we get
\[
    \exp(m_t) \geq c \sum_{t=0}^{t-1} \eta_t + 1 
\]
which yields the following logarithmic growth for the fixed learning rate schedule, i.e. $\eta_t =\eta$
\[
    m_t \geq \log( \eta c t + 1). 
\]
When te weights are not initialized at zero, there will be a moment $t_0$ where the margin will become positive and the same picture will hold.

We can then control the growth of the loss given by 
\[
    \cL(W_t) = \sum_x \log (1 + \exp(-(u_i-u_j)^T W_t e_x)) p(x) 
\]
which can be expressed in terms of the margins 
\[
    \cL(W_t) = \sum_x \log (1 + \exp(-m_t(x))) p(x). 
\]
Since $\log(1+e^{-x})$ is decreasing, we can directly install lower bound on $m_t(x)$ and get an upper bound on the loss  
\[
    \cL(W_t) \leq \sum_x \log (1 + \exp(-\log(\eta c_x (t-t_0(x))))) p(x) = \sum_x \log (1 + \frac{1}{\eta c_x (t-t_0(x))}) p(x) \leq \sum_x \frac{p(x)}{\eta c_x (t-t_0(x))}.
\]
When $W_0 = 0$, $t_0(x) \leq 0$, which implies the second part of the theorem.

\subsection{Multi-class - Proof of Theorem~\ref{thm:multi}}
\label{proof:ortho}

For the multi-class, consider $x\in[N]$ and assume that $f^*(x) = 1$.
Because the dynamics decouple, we can simplify notation with $w = W e_x \in \R^d$, and forget about the context variable $x$.
Let us denote
\[
  	p_w(j) \propto \exp(w^\top u_j), \qquad
	\ell(w) = -\log(p_w(1)).
\]
Consider the gradient flow dynamics
\[
  \frac{\diff w}{\diff t} = -\nabla \ell(w) = -\sum_{j\in[M]} p_w(j) (u_j - u_1). 
\]
Developing the probabilities, we get
\[
  \sum_{j\in[M]} \exp(w^\top u_j)\frac{\diff w}{\diff t} = \sum_{j\in[M]} \exp(w^\top u_j) (u_1 - u_j).
\] 
Let us denote $w_j = \scap{w}{u_j}$.
\[
  \sum_{j\in[M]} \exp(w_j)\frac{\diff w}{\diff t} = \sum_{j\in[M]} \exp(w_j) (u_1 - u_j).
\]
When the $(u_j)$ are orthonormal, we can project the last equation on the $(u_i)$, which leads to the following coupled differential equations
\[
	\sum_{j=1}^M \exp(w_j) \diff w_1 = \sum_{j=2}^M \exp(w_j) \diff t,\qquad\text{and}\qquad
	\sum_{j=1}^M \exp(w_j) \diff w_i = -\exp(w_i)\diff t \qquad \forall\, i\neq 1.
\]
We can rewrite it with the partition function $A(w) = \sum \exp(w_j)$,
\[
	\frac{A(w)}{A(w) - \exp(w_1)} \diff w_1 =  \diff t,\qquad\text{and}\qquad 
	A(w) \exp(-w_i) \diff w_i = -\diff t \qquad \forall\, i\neq 1.
\]
Subtracting any two instances of the coupling equations for $i,j\neq 1$, we get the following invariant
\begin{align*}
    & A(w) \paren{\exp(-w_i) \diff w_i - \exp(-w_j) \diff w_j} = 0
    \qquad\Leftrightarrow\qquad
    \exp(-w_i) \diff w_i - \exp(-w_j) \diff w_j = 0
    \\&\qquad\Leftrightarrow\qquad
    \exp(-w_i) - \exp(-w_j) = \exp(-u_i^\top w_0) - \exp(-u_j^\top w_0) =: c_{ij}
\end{align*}

The last invariant is found with
\[
    A(w)\diff w_1 =  (A(w) - \exp(w_1)) \diff t = \sum_{i > 1} \exp(w_i) \diff t = -\sum_{i > 1} A(w) \diff w_i
    \qquad\Leftrightarrow\qquad \sum_{i\in[M]} \diff w_i = 0.
\]
This is the transcription of the fact that the update of $w$ are in the span of the $(u_i - u_j)$ which is the orthogonal $(\sum u_i)^{\perp}$ when the $(u_i)$ are orthonormal. 

The first invariant allows us to characterize the partition function using only $w_1$ and the logit of the most probable incorrect class.  
Let $k = \argmin_{j \in \{2,..,M\} } \exp(-w_j)$, hence $c_{jk} \geq 0$ for $j \neq 1$, and 
\begin{align*}
    A(w) &= \exp(w_1) + \sum_{j\neq 1} \exp(w_j)
     = \exp(w_1) + \sum_{j\neq 1} \frac{1}{c_{jk} + \exp(-w_k)}
     \\&= \exp(w_1) + \sum_{j\neq 1} \frac{\exp(w_k)}{c_{jk}\exp(w_k) + 1}
     \\&= \exp(w_1) + (M_k+\theta_k) \exp(w_k)
\end{align*}
where
\[
     M_k = \card{\brace{k\neq 1|c_{jk}=0}} \geq 1, \qquad
     \theta_k \in [0, \card{\brace{k\neq 1|c_{jk} > 0}}] = [0, M - M_k],
\]
where we have used that when $c_{jk} > 0$, $\exp(w_k) / (c_k\exp(w_k) + 1) \leq \exp(w_k)$.

Note that for any $j\neq 1$, we can write the differential equations for the margin as
\[
    A(w)\diff (w_1 - w_j) = (A(w) - \exp(w_1) + \exp(w_j)) \diff t.
\]
In particular, for the tightest margin, we get
\[
  \frac{A(w)}{A(w) - \exp(w_1) + \exp(w_k)}
  = \frac{A(w)\exp(-w_k)}{A(w)\exp(-w_k) - \exp(w_1 - w_k) + 1}
  = \frac{\exp(w_1-w_k) + M_k + \theta_k}{M_k + \theta_k + 1}.
\]
Using the bounds on $\theta_k$, we get
\[
    \paren{\frac{\exp(w_1-w_k)}{M + 1} + \frac{M_k}{M_k + 1}} \diff (w_1 - w_k) 
    \leq \diff t
    \leq \paren{\frac{\exp(w_1-w_k)}{M_k + 1} + \frac{M}{M + 1}} \diff (w_1 - w_k) 
\]
Let us introduce constants to ease notations,
\[
    \paren{c_1\exp(w_1-w_k) + c_2} \diff (w_1 - w_k) 
    \leq \diff t
    \leq \paren{c_3\exp(w_1-w_k) + c_4} \diff (w_1 - w_k).
\]
We can integrate these inequalities
\[
    c_1\exp(w_1-w_k) + c_2(w_1 - w_k) + b_1
    \leq t
    \leq c_3\exp(w_1-w_k) + c_4 (w_1 - w_k) + b_2.
\]
This implies
\[
    w_1 - w_k = \log(t) (1 + o(1)).
\]
Let us denote $h(t) = \exp(w_1)$, using the first invariant, we get
\[
    \exp(w_1 - w_k) = \exp(w_1)(\exp(-w_j) + c_{kj}) = \exp(w_1 - w_j) + c_{kj}h(t) = c_t t(1 + o(t)) + c_{kj} h(t),
\]
for $c_t \in [1/c_3, 1/c_1]$ a bounded function.
We can characterize $h(t)$ with the last invariant
\[
    \sum_{i\in[M]} w_i =: C = o(1).
\]

The previous equations are solved with
\[
    w_i = \frac{-1}{M}\log(t) (1 + o(1)),
    \qquad w_1 = \frac{M-1}{M}\log(t) (1 + o(1)),
\]
which leads to $c_{kj} h(t) = c_{kj} t^{1-1/M} = o(t)$, and, since the $(u_i)$ are orthonormal,
\[
    w = \sum_{i\in[M]} \scap{w}{u_i} u_i = \sum_{i\in[M]} w_i u_i = \sum_{i\in[M]} \frac{\log(t)}{M} (u_1 - u_i).
\]
We can simplify the last equation by realizing that it is proportional to the projection of $u_1$ on the span of the $u_i - u_j$, which is also the span of the $u_1 - u_i$.
If we denote $\Pi$ the projection on this span, we have the existence of $(b_i)$ such that
\[
    \Pi(u_1) = \sum_{i>2} b_i (u_1 - u_i),
\]
and since $(\Pi(u_1) - u_1)^\top (u_i - u_j) = 0$ for all $i, j > 1$, we deduce $b_i = b_j = b$.

The value of $b$ can be computed explicitly, the triangle formed by $0$, $u_1$ and $\Pi(u_1)$ is both isosceles and rectangular, which leads to $\norm{\Pi(u_1) - u_1)} = \norm{\Pi(u_1)}$, $1 = \norm{u_i}^2 = \norm{\Pi(u_1) - u_1}^2 + \norm{\Pi(u_1)}^2$, hence $\norm{\Pi(u_1)} = 1 / \sqrt{2}$.
We also have $\norm{\Pi(u_1)} = \Pi(u_1)^\top u_1 = (M-1)b$, from which we deduce that the proportionality constant in Theorem \ref{thm:multi} is $(M - 1) / \sqrt{2}M$.

\subsubsection*{Indications for a proof in the case of gradient descent}
For gradient descent, we expect the same theorem to hold for two simple reasons, which, for simplicity, we do not formalize.
\begin{itemize}
    \item By convexity, there could only be one directional convergence for gradient descent (regardless of the initialization), and it has to be the same as the one for gradient flow.
    \item Because the level lines of loss are exponentially spaced, for any fixed learning rates, gradient descent will become a finer and finer approximation of gradient flow as $\norm{W}$ grows large.
\end{itemize}
Another way to proceed is to retake the previous arguments in the discrete setting.
For example, when initializing gradient descent with $W = 0$, one can check by recurrence that $\exp(w_i) = \exp(w_j)$ for all $i,j\neq 1$, this allows reducing the dynamics to a scalar evolution, which can be treated as in Theorem \ref{thm:ortho-bin}.

\section{Dynamics with two particles interfering}

In the setting of Theorem \ref{thm:two-flow}, Theorem \ref{thm:inter} plus a few lines of omitted derivations lead to the couplings
\begin{equation}
    \scap{\nabla \cL(W)}{(u_j - u_i) \otimes e_j}
    = \norm{u_2 - u_1}^2\paren{ \frac{p(i) \alpha}{1+\exp((u_i - u_j)^\top We_i)} - \frac{p(j)}{1+\exp((u_j - u_i)^\top We_j)}}.
\end{equation}
We remark that if $\alpha \leq 0$, there is no competition between the memory associations, the dynamics always advances in the cone $\R_+ \cdot e_1 + \R_+ \cdot e_2$, reinforcing both associations simultaneously.

Let us introduce the margin
\[
    m_j = (u_j - u_i)^\top W e_j = \scap{W}{(u_j-u_i)\otimes e_j}, \qquad\text{for}\qquad \brace{i,j} = \brace{1,2}.
\]
The previous equation can be rewritten as
\[
    \scap{\nabla \cL(W)}{(u_j - u_i) \otimes e_j}
    = \norm{u_2 - u_1}^2\paren{ \frac{p(i) \alpha}{1+\exp(m_i)} - \frac{p(j)}{1+\exp(m_j)}},
\]
For the gradient flow, it leads to the evolution
\begin{equation}
    \diff m_j = -\scap{\nabla \cL(W)}{(u_j-u_i)\otimes e_j}
    = \norm{u_2 - u_1}^2\paren{\frac{p(j)}{1+\exp(m_j)} - \frac{p(i) \alpha}{1+\exp(m_i)}}\diff t.
\end{equation}
Similarly, if we define the orthogonal vectors
\[
    f_1 = e_1 + e_2, \qquad
    f_2 = e_1 - e_2,
\]
as well as the statistics, 
\[
    \gamma_i = \frac{1}{2} (u_1 - u_2)^\top W f_i,
\]
we get $m_1 = \gamma_1 + \gamma_2$ and $m_2 = \gamma_2 - \gamma_1$, and $\gamma_1 = (m_1 - m_2)/2$, $\gamma_2 = (m_1 + m_2)/2$.
Hence,
\begin{align*}
    \frac{2}{\norm{u_2 - u_1}^2}\frac{\diff \gamma_2}{\diff t} 
    &= \frac{1}{\norm{u_2 - u_1}^2}\frac{\diff m_1 + \diff m_2}{\diff t}
    \\&= \paren{ \frac{p(1)}{1+\exp(m_1)} - \frac{p(2)\alpha}{1+\exp(m_2)}}
    + \paren{ \frac{p(2)}{1+\exp(m_2)} - \frac{p(1)\alpha}{1+\exp(m_1)}}
    \\&= \frac{p(1)(1-\alpha)}{1+\exp(m_1)} + \frac{p(2)(1-\alpha)}{1+\exp(m_2)}
    \\&= (1-\alpha)\paren{ \frac{p(1)}{1+\exp(\gamma_1  + \gamma_2)} + \frac{p(2)}{1+\exp(\gamma_2 - \gamma_1)}}
\end{align*}
Similarly
\begin{align*}
    \frac{2}{\norm{u_2 - u_1}^2}\frac{\diff \gamma_1}{\diff t} 
    &= \frac{1}{\norm{u_2 - u_1}^2}\frac{\diff m_1 - \diff m_2}{\diff t}
    \\&= \paren{ \frac{p(1)}{1+\exp(m_1)} - \frac{p(2)\alpha}{1+\exp(m_2)}}
    - \paren{ \frac{p(2)}{1+\exp(m_2)} - \frac{p(1)\alpha}{1+\exp(m_1)}}
    \\&= \frac{p(1)(1+\alpha)}{1+\exp(m_1)} - \frac{p(2)(1+\alpha)}{1+\exp(m_2)}
    \\&= (1+\alpha)\paren{ \frac{p(1)}{1+\exp(\gamma_1  + \gamma_2)} - \frac{p(2)}{1+\exp(\gamma_2 - \gamma_1)}}
\end{align*}
This explains the evolution in the main text.
We see that $\gamma_2$ will grow at least logarithmically, while $\gamma_1$ will be contained eventually because of the growth of $\gamma_2$.

\subsection{Proof of Theorem \ref{thm:two-flow}}
\label{app:two-tokens-interference}

We start by focusing on the gradient flow dynamics.
Recall that the evolution of the max-margin and orthogonal directions~$\gamma_2$ and~$\gamma_1$ is given by the following ODEs:
\begin{align}
    & \frac{\diff \gamma_1}{\diff ct} 
    =  \frac{(1+\alpha)p_1}{1 + \exp\paren{\gamma_2 + \gamma_1}} - \frac{(1+\alpha)p_2}{1 + \exp\paren{\gamma_2 - \gamma_1}} \label{eq:gamma1-ode}
   \\ &\frac{\diff \gamma_2}{\diff ct} 
    = \frac{(1-\alpha)p_1}{1 + \exp\paren{\gamma_2 + \gamma_1}} + \frac{(1-\alpha)p_2}{1 + \exp\paren{\gamma_2 - \gamma_1}}.
    \label{eq:gamma2-ode}
\end{align}

\subsubsection*{Lower bound in the max-margin direction $\gamma_2$}
From the evolution equation \eqref{eq:gamma2-ode} of the margin direction~$\gamma_2$ we deduce, using the fact that either $e^{\gamma_1} \leq 1$ or $e^{-\gamma_1} \leq 1$ for all $\gamma_1 \in \R$,
\[
   \frac{\diff \gamma_2}{\diff ct} 
    \geq  \frac{(1-\alpha)\min(p_1, p_2)}{1 + \exp\paren{\gamma_2}} = \frac{(1-\alpha)p_2}{1 + \exp\paren{\gamma_2}},
\]
since we have assumed without restriction that~$p_1 \geq p_2$.
We have solved the differential equation in Appendix \ref{proof:bin-ortho} (with a different constant). 
Using Grönwall's inequality, integrating this out leads to, when initialized at $W_0 = 0$,
\[
    \gamma_2 \geq \log(c_1 t+1) - h(c_1 t+1).
\]
where $c_1 = (1-\alpha)c p_2$ and $h$ as defined in Appendix \ref{proof:bin-ortho}, i.e. $h(x) = \tilde h(x) \log(x) / x$ with $\tilde h \in [1/2, 2]$.

\subsubsection*{Upper bound in the orthogonal direction $\gamma_1$}

Let us now consider $\gamma_1$.
First, note that whenever~$\gamma_1 \leq 0$, then we have~$\diff \gamma_1 \geq 0$, thanks to our assumption~$p_1 \geq p_2$.
In particular, with zero initialization~$W(0) = 0$, we then have~$\gamma_1(t) \geq \gamma_1(0) = 0$ throughout.

Let us know look for an upper bound.
For $\gamma_1$ to grow, we need $\diff\gamma_1 \geq 0$.
Denoting~$\bar \gamma := \log( \sqrt{p_1/p_2})$, this only possible when
\begin{align*}
    \frac{p_1}{1 + \exp\paren{\gamma_2 + \gamma_1}} - \frac{p_2}{1 + \exp\paren{\gamma_2 - \gamma_1}} \geq 0
    \quad&\Leftrightarrow\quad 
     p_1 - p_2 + p_1 \exp\paren{\gamma_2 - \gamma_1} \geq p_2 \exp\paren{\gamma_2 + \gamma_1}
    \\&\Leftrightarrow\quad 
     (p_1 - p_2) \exp(-\gamma_2) \geq p_2 \exp(\gamma_1) - p_1 \exp(-\gamma_1) \\
     &\Leftrightarrow\quad
     (p_1 - p_2) \exp(-\gamma_2) \geq \sqrt{p_1 p_2} (\exp(\gamma_1 - \bar \gamma) - \exp(-\gamma_1 + \bar \gamma)) \\
     &\Leftrightarrow\quad
     \sinh (\gamma_1 - \bar \gamma) \leq \frac{(p_1 - p_2) \exp(-\gamma_2)}{2\sqrt{p_1 p_2}}.
\end{align*}
We define
\begin{equation}
    \label{eq:cgamma}
    C(\gamma_2) := \frac{(p_1 - p_2) \exp(-\gamma_2)}{2\sqrt{p_1 p_2}}.
\end{equation}
We thus have
\[
    \sinh(\gamma_1 - \bar \gamma) \geq C(\gamma_2) \qquad\Leftrightarrow\qquad \diff \gamma_1 \leq 0.
\]
This implies that gradient flow will be bounded.
In particular, the lower bound on $\gamma_2$ gives us
\begin{align*}
    \exp(-\gamma_2) \leq \frac{\exp(h(c_1 t + 1))}{c_1 t+1} \leq \frac{\exp(2/e)}{c_1 t+1} 
\end{align*}
We conclude that, when $\gamma_1$ is initialized at zero, we have that $\diff \gamma_1(0) \geq 0$, and $\gamma_1$ will grow until reaching the point where $\diff \gamma_1 \leq 0$, which leads to a bound on $\gamma_1$ characterized by
\[
    \sinh(\gamma_1 - \bar \gamma) \leq C(\gamma_2) \leq \frac{p_1-p_2}{2\sqrt{p_1 p_2}} \frac{\exp(2/e)}{c_1 t+1}.
\]
This yields
\[
    \gamma_1(t) \leq \bar \gamma + \sinh^{-1} \paren{\frac{p_1-p_2}{\sqrt{p_1 p_2}} \frac{1.05}{c_1 t+1}}
    \leq \frac{1}{2} \log \paren{\frac{p_1}{p_2}} + \frac{p_1-p_2}{\sqrt{p_1 p_2}} \frac{1.05}{c_1 t+1}, 
\]
using that~$\sinh^{-1}(x) \leq x$ for~$x \geq 0$.

\subsubsection*{Upper bound in the max-margin direction $\gamma_2$}
We can upper bound $\gamma_2$ based on Equation \eqref{eq:gamma2-ode},
\[
   \frac{\diff \gamma_2}{\diff ct} 
    \leq  \frac{2(1-\alpha) \max(p_1, p_2)}{1 + \min(\exp(\gamma_1), \exp(-\gamma_1)) \exp\paren{\gamma_2}}
    = \frac{2(1-\alpha) p_1}{1 + \exp(-\gamma_1) \exp\paren{\gamma_2}}.
\]
We have seen that
\begin{align*}
    \exp(\gamma_1) 
    &\leq \sqrt{\frac{p_1}{p_2}}\exp(\sinh^{-1}(\frac{p_1-p_2}{p_2} \frac{1.05}{c_1 t+1}))
    = \sqrt{\frac{p_1}{p_2}} \paren{\frac{p_1-p_2}{p_2}\frac{1.05}{c_1 t+1} + \sqrt{\frac{(p_1-p_2)^2}{p_2^2}\frac{1.05}{(c_1 t+1)^2} + 1}}
    \\&\leq \sqrt{\frac{p_1}{p_2}}\paren{1.05\frac{p_1-p_2}{p_2} + \sqrt{1.05\frac{(p_1-p_2)^2}{p_2^2} + 1}}
    \leq \frac{p_1}{p_2}\paren{(1.05  + \sqrt{2.05})\frac{p_1 - p_2}{p_2} + \sqrt{2.05}}
    \leq 4 \paren{\frac{p_1}{p_2}}^2.
\end{align*}
We deduce that
\[
   \frac{\diff \gamma_2}{\diff ct} 
   \leq  \frac{2(1-\alpha) p_1}{1 +\frac{1}{4} \paren{\frac{p_2}{p_1}}^2 \exp\paren{\gamma_2}}
   \leq  \frac{8(1-\alpha) p_1^3 / p_2^2}{1 + \exp\paren{\gamma_2}}.
\]
This allows us to conclude that $\gamma_2$ does not grow faster than logarithmically.
\[
    \gamma_2 \leq \log(c_2 t+1) - h(c_2 t+1).
\]
with $c_2 = 8 c (1 - \alpha) p_1^3 / p_2^2$.
Using the intermediate value theorem, we deduce the form of $\gamma_2$ given in the theorem.

\subsubsection*{Lower bound in the orthogonal direction $\gamma_1$}

If $\gamma_1$ was initialized such that $\diff \gamma_1 \leq 0$, we would have that $\gamma_1$ would decrease until reaching the point found in the upper bound for $\gamma_1$.
The difficulty consists in showing that $\gamma_1$ increases fast enough toward $\bar \gamma$.
Retaking the derivations made to characterize the sign of $\diff \gamma_1$, we can rewrite the evolution equation as
\begin{align*}
    \frac{\diff \gamma_1}{\diff ct} 
    &=  (1+\alpha)\frac{p_1 - p_2 + 2\sqrt{p_1p_2}\exp(\gamma_2)\sinh(\bar\gamma - \gamma_1)}{(1 + \exp\paren{\gamma_2 + \gamma_1})(1 + \exp\paren{\gamma_2 - \gamma_1})}
    \\&=  (1+\alpha)\frac{p_1 - p_2 + 2\sqrt{p_1p_2}\tilde c_t t \sinh(\bar\gamma - \gamma_1)}{(1 + \tilde c_t t\exp\paren{\gamma_1})(1 + \tilde c_t t\exp\paren{- \gamma_1})},
\end{align*}
where $\tilde c_t / c_t \in [\exp(e/2), 1]$ is found with the intermediate value theorem, and we have used that
\[
   \exp(\gamma_2) = c_t t\exp(-h(c_t t + 1)) \in c_t t \cdot [\exp(e / 2), 1].
\]
We can lower bound the growth of $\gamma_1$ when $\gamma_1 \leq \bar\gamma$, which implies $\diff \gamma_1 \geq 0$.
Using that $\gamma_1$ is bounded, we get the existence of a constant $c_3$ such that
\[
    \frac{\diff \gamma_1}{\diff ct} 
	\geq  (1+\alpha)\frac{p_1 - p_2 + 2\sqrt{p_1p_2}\tilde c_t t\sinh(\bar\gamma - \gamma_1)}{(1 + c_3 t)^2}
	\geq  (1+\alpha)\frac{p_1 - p_2}{(1 + c_3 t)^2}.
\]
This leads to a growth of $\gamma_1$ in $O(1/t)$, from which we deduce that
\[
  	\gamma_1 \geq \bar\gamma + O(1/t),
\]
which ends the characterization of the dynamics for gradient flow.

\subsubsection*{Gradient Descent}

The dynamics of~$\gamma_1$ and~$\gamma_2$ for gradient descent with a step-size~$\eta$ are given by
\begin{align*}
    \gamma_1(t + 1) = \gamma_1(t) + \eta c \Delta \gamma_1, \qquad \gamma_2(t + 1) = \gamma_2(t) + \eta c \Delta \gamma_2,
\end{align*}
with
\begin{align}
    & \Delta \gamma_1 
    =  \frac{(1+\alpha)p_1}{1 + \exp\paren{\gamma_2 + \gamma_1}} - \frac{(1+\alpha)p_2}{1 + \exp\paren{\gamma_2 - \gamma_1}} \label{eq:gamma1-delta}
   \\ & \Delta \gamma_2
    = \frac{(1-\alpha)p_1}{1 + \exp\paren{\gamma_2 + \gamma_1}} + \frac{(1-\alpha)p_2}{1 + \exp\paren{\gamma_2 - \gamma_1}}.
    \label{eq:gamma2-delta}
\end{align}

Similar to gradient flow, we can lower bound the update equation of $\gamma_2$ for descent, with $c_1 = (1-\alpha) c p_2$
\[
   \gamma_2(t+1) - \gamma_2(t) \geq \frac{\eta c_1}{1 + \exp\paren{\gamma_2}}.
\]
Since we expect logarithmic growth from the study of flow, we want to study $\exp(\gamma_2(t))$.
In particular 
\[
    \exp(\gamma_2(t+1)) \geq \exp\left( \frac{\eta c_1}{1 + \exp\paren{\gamma_2}} \right) \exp(\gamma_2(t))
\]
Using $e^x \geq 1+x$, we furthermore get
\[
    \exp(\gamma_2(t+1)) - \exp(\gamma_2(t)) \geq \eta c_1 \cdot\frac{\exp(\gamma_2(t))}{1 + \exp\paren{\gamma_2(t)}}.
\]
Since $\gamma_2$ is always non-negative we have that $\exp(\gamma_2)/(1+\exp(\gamma_2))\geq 1/2$ hence we get a recursion 
\begin{align}
    \exp(\gamma_2(t+1)) - \exp(\gamma_2(t)) \geq \eta c_1 / 2,
\end{align}
Using a telescopic sum, we get the desired lower bound $\gamma_2(t) \geq \log(\eta c_1  t / 2 + 1)$.

Let us know focus on~$\gamma_1$.
In comparison to gradient flow, $\gamma_1$ can grow large because of potentially large steps taken from values where $\Delta \gamma_1$ is positive.
Similar to the gradient flow case, we have that~$\Delta \gamma_1 \leq 0$ if and only if
\[
    \sinh(\gamma_1 - \bar \gamma) \geq C(\gamma_2),
\]
where~$C(\gamma_2)$ is given in~\eqref{eq:cgamma}.
Now consider a time~$t$.
If~$\sinh(\gamma_1(t) - \bar \gamma) \leq C(\gamma_2(t))$, we have
\begin{align*}
    \gamma_1(t) 
    &\leq \gamma_1(t+1) \leq \gamma_1(t) + \eta c \Delta \gamma_1 \leq \bar \gamma + \sinh^{-1}(C(\gamma_2(t))) + \eta c(1 - \alpha) p_1 
    \\& \leq \bar \gamma + C(\gamma_2(0)) + \eta c(1 - \alpha) p_1
    \leq \bar \gamma + \frac{p_1}{2p_2} + \eta c(1 - \alpha) p_1 =: \gamma_{\max}.
\end{align*}
If~$\sinh(\gamma_1(t) - \bar \gamma) \geq C(\gamma_2(t))$, then~$\gamma_1(t+1) \leq \gamma_1(t)$, and
\[
\gamma_1(t+1) \geq \gamma_1(t) + \eta c \Delta \gamma_1 \geq \bar \gamma -\eta c (1 - \alpha) p_2 =: \gamma_{\min},
\]
where~$\gamma_1(t) \geq \bar \gamma$ follows from~$\sinh(\gamma_1(t) - \bar \gamma) \geq 0$.

By induction, we then have that~$\gamma_1(t) \in [\min(0, \gamma_{\min}), \gamma_{\max}]$ for all~$t$ (assuming~$\gamma_1(0) = 0$).

For simplicity, we skip the upper bound on $\gamma_2$, as well as the convergence of $\gamma_1$ towards $\bar\gamma$.

\subsubsection*{Indications to prove that $\gamma_1$ converges to $\bar \gamma$}
Note that in the case of gradient descent with large learning rates, $\gamma_1$ might be oscillating around $\bar \gamma$.
This case does not happen in gradient flow and requires extra derivations to handle it.
Using the fact that~$\gamma_2(t) \geq \log(\eta c t + 1)$, we get
\[
C(\gamma_2(t)) \leq \frac{p_1 - p_2}{2\sqrt{p_1 p_2}} \frac{1}{\eta c_3 t + 1}.
\]
If~$\sinh(\gamma_1(t) - \bar \gamma) \leq C(\gamma_2(t))$, we then have
\begin{align*}
\gamma_1(t) &\leq \gamma_1(t+1) = \gamma_1(t) + \eta c \Delta \gamma_1 \\
    &\leq \bar \gamma + \frac{p_1 - p_2}{2\sqrt{p_1 p_2}} \frac{1}{\eta c_3 t  + 1} + \frac{\eta c(1 - \alpha) p_1}{ 1 + \exp(\gamma_{\min}) \eta c_3 t} = \bar \gamma + O(1/t)
\end{align*}
If~$\sinh(\gamma_1(t) - \bar \gamma) \geq C(\gamma_2(t))$, then
\begin{align*}
    \gamma_1(t) &\geq \gamma_1(t+1) = \gamma_1(t) + \eta c \Delta \gamma_1 \\
        &\geq \bar \gamma - \frac{\eta c (1 - \alpha) p_2}{1 + \exp(-\gamma_{\max}) \eta c_3 t} = \bar \gamma + O(1/t).
\end{align*}
This ensures that when the dynamics are in an oscillating regime, the bound on~$|\gamma_1(t) - \bar \gamma|$ will decrease as~$O(1/t)$, thus inducing faster progress towards perfect accuracy than as guaranteed by the looser bound~$[\gamma_{\min}, \gamma_{\max}]$.

\subsection{Loss Spike}
\label{proof:spike}

Proposition \ref{thm:spike} follows from
\[
    \cL(W_1) \geq p_2 \ell(W_1; 2,2) = p_2 \log(1 + \exp(-m_2)) \geq -p_2 m_2.
\]
In particular, when initialized at zero, after one gradient update
\[
    m_2 = \eta (p_2 - \alpha p_1).
\]

\section{Transformer experiments}
\label{sub:appx_transformer}

In this section, we provide more details on the setup for the transformer experiments in Section~\ref{sub:transformer}.

We follow~\citet{bietti2023birth} and consider a simplified two-layer Transformer architecture trained on a simple in-context learning task.
The task consists of sequences of tokens~$z_{1:T} \in [N]^T$, where any occurrence of a so-called \emph{trigger} token~$q \in Q$ is followed by the same \emph{output} token~$o_q$, but where~$o_q$ is resampled uniformly across different sequences.
The tokens following all non-trigger tokens are randomly sampled from a sequence-independent Markov model (namely, a character-level bigram model estimated from Shakespeare text data).

We focus on the prediction of the output tokens~$o_q$ given a sequence~$[z_1, \ldots, q, o_q, \ldots, q]$, where we assume~$q$ has appeared at least once before the last token.
Correctly predicting the token~$o_q$ then requires finding previous occurrences of~$q$ in the input sequence and copying the token just after it.
\citet{bietti2023birth} show that this task can be solved with a two-layer transformer with no feed-forward blocks, and all layers fixed at random except three trained matrices, by implementing an ``induction head'' mechanism~\citep{elhage2021mathematical,olsson2022context}.
The three trained matrices were found to behave as associative memories, each with different sets of embeddings, as we now detail:
\begin{itemize}
    \item $W_K^1$ (first layer key-query matrix), which implements a previous token lookup, satisfying
    \[\arg\max_j \langle p_j, W_K^1 p_t \rangle = t - 1,\]
    where~$p_t$ are positional embeddings;
    \item $W_K^2$ (second layer key-query matrix), which implements lookup of the previous trigger that matches the current token, with
     \[\arg\max_j \langle e_j, W_K^2 \phi_k(e_i) \rangle = i,\]
     where~$e_i$ are input token embeddings, and~$\phi_k(e_i) = W_O^1 W_V^1 e_i$ is a remapping of the input embeddings by the first attention head;
    \item $W_O^2$ (second layer output matrix), which implements a copy of the output token into the unembedding space, with
    \[
    \arg\max_j \langle u_j, W_O^2 \phi_o(e_i) \rangle = i,
    \]
    where~$u_j$ are output embeddings, and~$\phi_o(e_i) = W_V^2 e_i$ is a remapping of input embeddings by the (random) second value matrix.
\end{itemize}
We may then define the $W_O^2$ margins (the ones for~$W_K^{1/2}$ are defined analogously):
\[
m_i = \langle u_i, W_O^2 \phi(e_i) \rangle - \max_{j \ne i} \langle u_j, W_O^2 \phi(e_i) \rangle.
\]

As explained in~\citep{bietti2023birth}, we note that input embeddings to each matrix are often sums/superpositions of embeddings, some of which are typically noise that gets filtered out during training. For instance, training~$W_O^2$ alone may recover the desired associations in high-dimension, even though its input at initialization is an average over all tokens in the sequence, due to the initially flat attention pattern.
Our training setup is the following: 
we consider full-batch gradient descent on a dataset of 16\,384 sequences of length 256 generated from the model described above with~$N=64$ tokens.
The loss considers only predictions on tokens~$o_q$, ignoring the very first occurrence since it is not predictable from context.

\end{document}